%% file: paper.tex
\documentclass[sigconf]{acmart}
\copyrightyear{2023} 
\acmYear{2023} 
\renewcommand\footnotetextcopyrightpermission[1]{} %

\usepackage{amsmath,amsfonts}
\usepackage{graphicx}
\DeclareMathOperator*{\argmax}{arg\,max}

\usepackage{graphicx}
\usepackage{hyperref}
\usepackage{url}
\usepackage{float}
\usepackage{setspace}
\usepackage{harpoon}
\usepackage{multicol}
\usepackage{multirow}
\usepackage{caption}
\usepackage{hhline}
\usepackage[labelformat=simple]{subcaption}
\usepackage{enumitem}
\usepackage{url}
\usepackage{lipsum}

\newcommand*{\origrightarrow}{}

\renewcommand*{\textrightarrow}{\fontfamily{cmr}\selectfont\origrightarrow}

\newcommand{\etal}{\textit{et al.}}

\AtBeginDocument{%
  \providecommand\BibTeX{{%
    \normalfont B\kern-0.5em{\scshape i\kern-0.25em b}\kern-0.8em\TeX}}}

\acmConference[ACM Multimedia 2023]{ACM Multimedia 2023}{October 29, – November 3, 2023}{Ottawa, Canada}

\begin{document}

\title{Finding Meaningful Distributions of ML Black-boxes\\under Forensic Investigation}
\author{Jiyi Zhang, Han Fang, Hwee Kuan Lee, Ee-Chien Chang}
\affiliation{%
  \institution{School of Computing\\National University of Singapore}
  \country{Singapore}
}

\input{abstract.tex}




\maketitle
\pagestyle{plain}
\makeatletter
\renewcommand\@formatdoi[1]{\ignorespaces}
\makeatother

\input{intro}
\input{background}

\input{approach}
\input{construction}

\input{eval}

\input{countermeasure}
\input{conclusion}

\bibliographystyle{ACM-Reference-Format}
\bibliography{paper}

\end{document}

%% file: abstract.tex
\begin{abstract}

Given a poorly documented neural network model, we take the perspective of a forensic investigator who wants to find out the model’s data domain (e.g. whether on face images or traffic signs). Although existing methods such as membership inference and model inversion can be used to uncover some information about an unknown model, they still require knowledge of the data domain to start with. In this paper, we propose solving this problem by leveraging on comprehensive corpus such as ImageNet to select a meaningful distribution that is close to the original training distribution and leads to high performance in follow-up investigations. 
The corpus comprises two components, a large dataset of samples and meta information such as hierarchical structure and textual information on the samples. Our goal is to select a set of samples from the corpus for the given model. The core of our method is an objective function that considers two criteria on the selected samples: the model functional properties (derived from the dataset), and semantics (derived from the metadata). We also give an algorithm to efficiently search the large space of all possible subsets w.r.t. the objective function.
Experimentation results show that the proposed method is effective. For example, cloning a given model (originally trained with CIFAR-10) by using Caltech 101 can achieve 45.5\% accuracy. By using datasets selected by our method, the accuracy is improved to 72.0\%.

\end{abstract}

%% file: intro.tex
\section{Introduction}

Machine learning models, in particular, deep neural networks have achieved excellent performance in many tasks. The development of easy-to-use training tools and widely available datasets further accelerate  their adoption. While the wide adoption is encouraging,  it would not be surprising to see involvement of ML models in illegal activities.  

Here, we take the perspective of forensic investigators, who want to investigate some suspicious ML models and determine the purpose of the models. These ML models could be in the form of black-boxes and embedded in some apprehended devices with internal mechanisms hidden, for instance, the models are embedded in FPGA, ASIC, or Trusted Execution Environment such as Intel SGX~\cite{10.1145/2995306.2995307}. They could also be in the form of remote services that receive and send inputs/predictions over the Internet. 

Although we adopt a black-box threat model,
in practice, in the white-box setting (where investigators can examine parameters and intermediate states), due to lack of transparency/explainability, limited information can be obtained. Therefore, such investigation is also applicable to poorly documented white-box models, for example, model files stored in Google Drive with format such as `.pth' or `.ckpt'.

Methods such as membership inference~\cite{DBLP:conf/ccs/JiaSBZG19,DBLP:journals/tdp/RahmanRLM18,DBLP:conf/ccs/ChenYZF20,DBLP:journals/corr/abs-2007-15528,DBLP:journals/corr/ShokriSS16,DBLP:conf/ndss/Salem0HBF019}, model cloning~\cite{DBLP:conf/uss/TramerZJRR16,DBLP:conf/ccs/PapernotMGJCS17,DBLP:series/lncs/OhSF19,DBLP:conf/sp/WangG18,DBLP:conf/kdd/LowdM05,DBLP:conf/fat/MilliSDH19,DBLP:conf/cvpr/OrekondySF19} and model inversion~\cite{DBLP:conf/ccs/FredriksonJR15,DBLP:journals/tnn/LeeK94,DBLP:journals/tnn/LuKN99,DBLP:journals/corr/MahendranV14,DBLP:conf/cvpr/DosovitskiyB16,DBLP:conf/nips/DosovitskiyB16,DBLP:conf/aistats/NashKW19,DBLP:conf/ccs/YangZCL19} can be deployed to provide useful information of an unknown model. However, most of these methods require some knowledge about the data domain of the model to start from, for instance, whether they are face images, X-ray images, etc.  These methods either become ineffective or have decreased performance if such information is inaccurate. 

In this paper, we work on a new task which fills in the above-mentioned gap by determining the data domain of model with black-box and hard-label only access. 
We do not aim to find the exact original data distribution nor the training set that the model was trained on. Instead, we are  contended with some meaningful datasets that could lead to high performance from the follow-up investigation.  

Due to the domain's high dimensionality, there could be many drastically different training sets that lead to the target model. 
We leverage on comprehensive corpus gathered by the research community as a starting point, based on the assumption that the black-box models are trained on meaningful images with similar semantics captured  in the corpus. 
\\ \\
\noindent
{\em Problem Statement.}\ \
Given a black-box model and access to a comprehensive corpus, the investigator wants to find a candidate distribution which is close to the original training data distribution of the model.

A corpus contains a large dataset and metadata which links data points according to their semantics. The investigator wants to identify whether the corpus contains data points which are similar to the target model's training data. Then the investigator extracts a subset from the corpus to form a candidate distribution as the outcome of the investigation.

In our approach, we use an objective function which is a weighted sum of two terms to find a good distribution.
The first term focuses on the black-box model's response to input data and covers the functional aspect of the target model.
The second term makes use of the metadata, in particular, the hierarchical structure, to measure how semantically meaningful the candidate distribution is. 
As the search space is extremely large,
We propose a heuristic to efficiently search for a good candidate distribution.

Experimentation results demonstrate that the proposed approach can effectively find candidate datasets that are close to the original input distribution and boost the performance of further investigations. For example, when conducting model cloning on a model originally trained on a subset of CIFAR-10 with a fixed small number of samples, using a similar dataset (Caltech101) can achieve 45.5\% testing accuracy and using our selected dataset can achieve 72.0\%. 
We also conducted experiments on two undocumented models downloaded from public domain. Using our method, we suspect that the first one is trained on ImageNet and classifies the 1,000 classes in  ImageNet.  For the second one, we suspect that it is trained on three classes of plants. Further analysis suggests that it may be trained on some leaves that look like bean leaves.
\\ \\
\noindent
{\bf Contributions.}
\begin{enumerate}[leftmargin=*,noitemsep]
    \item We highlight that determining an accurate input distribution of a given model is an important first step for machine learning forensic investigations.
    \item We propose a method to find a dataset that is close to the training distribution of a given classifier. This method uses background information supplied by a hierarchical corpus. It has two main components: an objective function that evaluates whether a candidate dataset is good and an algorithm to search from all subsets of the corpus.
    \item We evaluate the effectiveness of our approach under several settings: (a) Find the input data distribution for classifiers with known ground truth,
    (b) Find the input data distribution for unknown models hosted in model hubs,
    (c) Use the datasets obtained through the proposed method for follow-up investigation, in particular, model cloning.
\end{enumerate}

%% file: background.tex
\section{Background and Related Works}

The problem we are trying to solve is related but different from the goal of model inversion and membership inference. This section gives a brief overview of these two works. In addition, the main purpose of our method is to find a suitable dataset for follow-up investigations, such as model cloning. This section also includes an introduction to model cloning.

\subsection{Model Inversion}
Model inversion is a method which aims to extract information related to the input data of a model. 
Most model inversion approaches focus on reconstructing an input image from the soft predictions. In such cases, given a neural network classification model ${\mathcal M}:\mathcal{X} \rightarrow \mathbb{R}^n$ trained on a data distribution ${\mathcal D}$, the goal of model inversion is to create a model $\widehat{{\mathcal M}}:\mathbb{R}^n \rightarrow \mathcal{X}$ such that for ${\textbf x}$ sampled from $ {\mathcal D}$,  the new model is able to reconstruct ${\textbf x}$ from the target classifier's output ${\mathcal M}({\textbf x})$, i.e. $\widehat{{\mathcal M}}({\mathcal M}({\textbf x})) = {\textbf x}$.

Model inversion can be categorized into gradient-based methods~\cite{DBLP:conf/ccs/FredriksonJR15,DBLP:journals/tnn/LeeK94,DBLP:journals/tnn/LuKN99,DBLP:journals/corr/MahendranV14} which invert a model using some optimization functions, and training-based methods~\cite{DBLP:conf/cvpr/DosovitskiyB16,DBLP:conf/nips/DosovitskiyB16,DBLP:conf/aistats/NashKW19,DBLP:conf/ccs/YangZCL19} which learn new substitute models as the inverse of original target models. Both categories require access to soft predictions of the target models. Some methods also need white-box access to exploit backpropagation.

If only hard-label access is permitted, the problem can also be further reduced to class inversion which creates representative samples for each class. Though model inversion is able to reconstruct a single input or create representative samples for a class, background information about the training distribution is still required.

\subsection{Membership Inference}
Given a model $\mathcal{M}$ and a particular sample ${\textbf x}$, the purpose of membership inference is to find out whether this sample ${\textbf x}$ is included in the training dataset $\mathcal{D}$ of the model $\mathcal{M}$. There are different versions of membership inference with different settings~\cite{DBLP:conf/ccs/JiaSBZG19,DBLP:journals/tdp/RahmanRLM18,DBLP:conf/ccs/ChenYZF20,DBLP:journals/corr/abs-2007-15528,DBLP:journals/corr/ShokriSS16,DBLP:conf/ndss/Salem0HBF019}. Most existing works assume the investigator has some knowledge about the architecture of the target model. In addition, the investigator also needs some background about the distribution of the target model's training dataset. Such background knowledge may be in the form of some samples from same distribution but not included in the training dataset.

\subsection{Model Cloning}
\label{sec:model_cloning}
Given a black-box neural network classification model $\mathcal{M}:\mathcal{X} \rightarrow \mathbb{R}^n$ trained on a data distribution ${\mathcal D}$, the goal of model cloning is to create a model $\widetilde{\mathcal M}:\mathcal{X} \rightarrow \mathbb{R}^n$ such that for  most  ${\textbf x}$ sampled from $ {\mathcal D}$,  the new model produces the same prediction result as the original target model, i.e. $\widetilde{\mathcal M}({\textbf x}) = \argmax \mathcal{M}({\textbf x})$.

There are extensive studies on model cloning. Tramer~\etal~proposed cloning single layer logistic regression models using equation solving method and cloning decision tree models using path finding method~\cite{DBLP:conf/uss/TramerZJRR16}. Their methods do not require the usage of natural samples but assume that the investigator knows about architecture and training procedures of the target model. Soft predictions (a.k.a confidence scores) are also required. Papernot \etal~demonstrated the effectiveness of model cloning on more complicated models~\cite{DBLP:conf/ccs/PapernotMGJCS17}. Their method does not require the knowledge about target model's architecture or training details. They use synthesized natural samples from the same distribution as the target model's training data and applied a fixed training strategy for cloning. Only hard labels are required. Methods with similar goals were also explored in many other literatures~\cite{DBLP:series/lncs/OhSF19,DBLP:conf/sp/WangG18,DBLP:conf/kdd/LowdM05,DBLP:conf/fat/MilliSDH19,DBLP:conf/cvpr/OrekondySF19}.

%% file: approach.tex
\section{Problem Formulation}
The investigator has hard-label black-box access to a model ${\mathcal M}$ with $n$ classes. That is, the investigators can adaptively submit inputs to the black-box, and receive the output in the form of one-hot vectors. 
For the given ${\mathcal M}$, let us write ${\mathcal M}({\bf x})$ as the array of soft labels of all classes on input ${\bf x}$ and ${\mathcal M_i}({\bf x})$ as the soft label for the $i$-th class on input ${\bf x}$.

The investigator also has a corpus $\mathcal{C} = \langle\mathcal{D},\mathcal{T}\rangle$, which consists of a dataset $\mathcal{D}$ and the associated metadata $\mathcal{T}$. For example, $\mathcal{D}$ is the set of images in ImageNet, and the metadata $\mathcal{T}$ contains the textual description of each class and describes the hierarchical relationship among the classes.

\subsection{Output of Investigation}

The investigator wants to select a set of samples from dataset $\mathcal{D}$. For each class $i$,
we define $\omega_{i}$ as the set of indices of the selected samples. For example, if  $\omega_{i} = \{i_1, i_2,...,i_k\}$, then $i_1 \text{-th}, i_2 \text{-th},...,i_k \text{-th}$  samples in $\mathcal{D}$ are being selected. In addition, we denote the set of images with indices in $\omega'$ as $\mathcal{D_{\omega'}}$. 

In summary, the output of the investigation is the set of selection parameters for $n$-classes, denoted as $\omega=\langle \omega_1, \omega_2, \ldots, \omega_n\rangle$.

The $n$-class dataset selected by the selection parameters is denoted as ${\mathcal D}_{\omega}=\langle {\mathcal D}_{{\omega}_1}, {\mathcal D}_{{\omega}_2}, \ldots, {\mathcal D}_{{\omega}_n} \rangle$.

\subsection{Objectives}
For each class $i$, the selection parameter $\omega_i$ should be able to select $D_{\omega_i}$ that meets four objectives: responsiveness, transformation invariance, inclusiveness and common sense. 

We give details of the objective function in Section~\ref{sec:obj}. Here, we explain the intuitions of the objectives.
We divide these objectives into two categories: functional requirement and semantic requirement. Functional requirement is derived from the dataset part of the corpus while semantic requirement is derived from metadata.  

\subsubsection{Functional Requirement}\ \
A good candidate distribution should receive correct responses from the target model. For example, most samples from the candidate distribution of class $i$ should be classified as class $i$ by the target model. In addition, the model should have some kind of transformation invariance on the candidate distribution.
We break down the functional requirement into responsiveness and transformation invariance.
\\ \\
\noindent
{\em Responsiveness.}\ \
Responsiveness looks at the outputs of the target model when a dataset is fed into the model. 
It measures the consistency of the model predictions when taking in samples from a given distribution as inputs. For example, the responsiveness is considered low when images of a same class object causes different predictions.
Give a target model, for each class, we may set the responsiveness score as the percentage of the samples of same ground truth which get classified into that class by the target model.
\\ \\
{\em Transformation Invariance.}\ \
Measuring responsiveness of single data points may not be sufficient in all scenarios. A trained model should have certain transform-invariance to be useful in real life. It should be able to tolerate some noise, translation, rotation, perspective projection, etc. For data distributions which the model is not trained on, such invariance usually does not exist. Therefore, instead of using single data points for analysis, we can draw samples from a data point's neighborhood and measure their responsiveness to make sure the results are consistent.

\subsubsection{Semantic Requirement}\ \
    A good candidate distribution should represent a semantically meaningful class of things that exist in real life such as animals, tools, cars and X-ray images, instead of a set of unrelated images.
We break down semantic requirement into inclusiveness and common sense.
\\ \\
\noindent
{\em Inclusiveness.}\ \
When searching for a candidate distribution for a target model, we would like to select a distribution that is as general as possible. For example, given a target model trained on face data, images of a single person will also meet the objectives of responsiveness and transformation invariance. However, the correct distribution is more likely to be `human faces' instead of `faces of person A'.
\newpage
\noindent
{\em Common Sense.}\ \
The distribution selected for the target model should be a class of objects/characters/styles or a meaningful combination of above classes. In other words, it should satisfy some common sense. This objective is to filter out `noise' in the candidate distribution. For example, the distribution of `airliner + warplane + wing + shark' could be filtered. The distribution becomes more meaningful with shark removed and the rest re-grouped as `aircraft'. This step requires the background knowledge provided by the hierarchical corpus to decide which kind of combinations is more meaningful.

\section{The Proposed Approach: Using ImageNet as Corpus}
\label{sec:approach}
The metadata in ImageNet is presented as a rooted tree which links nodes of data points together.
Each node in ImageNet tree corresponds to a data class and is labelled (e.g. `airplane').  An internal node in the tree is the union of its children (see the tree in Figure~\ref{fig:overview}). 

There are two components  in the approach: (1) An objective function that measures whether a dataset $\mathcal{D}_{\omega_i}$ is a good fit w.r.t. the class $i$ in the target model and (2) an algorithm to search from all subsets of the corpus and maximize the score of objective function.
\input{figures/bigfigure.tex}
\subsection{Objective Function}
\label{sec:obj}
We investigate the data domain of a target model class by class. For each class $i$, we define the objective function as weighted sum of two terms: functional score and semantic score. That is:
$$
\alpha \cdot \mathcal{L}_{\mathcal{M}_{i},\mathcal{D}}({\omega_i}) + (1-\alpha)\cdot \mathcal{S}_{\mathcal T}({\omega_i})
$$
where $\alpha\in(0,1)$ is the weight. 

The functional score $\mathcal{L}_{\mathcal{M}_{i},\mathcal{D}}({\omega_i})$ combines the goals of responsiveness and transformation invariance by measuring the responsiveness of the model when the dataset $\mathcal{D}_{\omega_i}$ is fed into the model with respect to some transformations.
The semantic score $S_{\mathcal T}({\omega_i})$ combines the goals of inclusiveness and common sense and measures how semantically meaningful the selection $\omega_i$ is.  

Note that input of functional score is a dataset, whereas input of semantic score is the selection parameter.

\subsubsection{Functional Score}\ \
\label{sec:similarity_score}
We define the functional score of a class $i$ in the model with regard to $\mathcal{D}_{\omega_i}$ in the following way:
\begin{enumerate}
    \item Expand $\mathcal{D}_{\omega_i}$ by applying uniform random perturbation and a set of transformations including translation, rotation and perspective projection to each data point in $\mathcal{D}_{\omega_i}$. We denote this expanded set as $\mathcal{E}(\mathcal{D}_{\omega_i})$.
    \item The functional score w.r.t $i$-th class is the percentage of the samples in $\mathcal{E}(\mathcal{D}_{\omega_i})$ which get classified into it, that is:$$\mathcal{L}_{\mathcal{M}_i,\mathcal{D}}({\omega_i}) = \frac{|\{{\textbf x} \in \mathcal{E}(\mathcal{D}_{\omega_i}): \argmax(\mathcal{M}({\textbf x}))=i\}|}{|\mathcal{E}(\mathcal{D}_{\omega_i})|}$$ 
\end{enumerate}

To see why this functional score makes sense, we can look at two aspects:
\begin{itemize}
    \item Firstly, a sensible model should be trained to classify a group of similar objects into a fixed class with high accuracy. Therefore, most samples from the same or very similar distributions should be classified into the same class.
    \item  Secondly, a trained model should have certain transform-invariance to be useful in real life. It should be able to tolerate some noise, translation, rotation, perspective projection, etc. For data distributions which the model is not trained on, such invariance usually does not exist.
\end{itemize}

%% file: figures/bigfigure.tex
\begin{figure}[H]
    \centering
    \includegraphics[width=\linewidth]{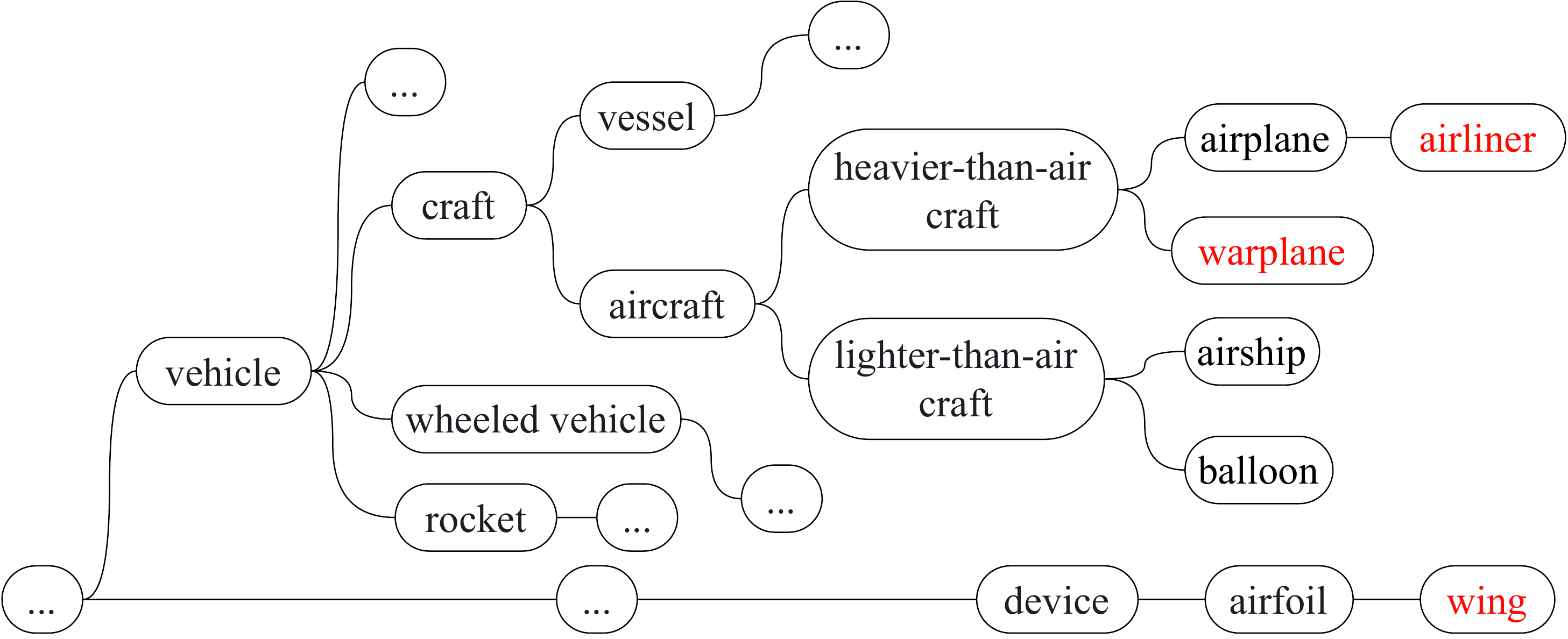}
    \caption{The proposed approach relies on a hierarchical corpus such as ImageNet. The hierarchical corpus is a collection of sets organized as a rooted tree, where each node corresponds to a set and is labelled (e.g. ``airplane''). An internal node in the tree is the union of its children. The search algorithm uses both the outputs of the target model and the metadata information provided by the hierarchical corpus to give a most suitable distribution.}
    \label{fig:overview}
\end{figure}

%% file: construction.tex
\subsubsection{Semantic Score}\ \
\label{sec:semantic_score}
We define the semantic score for a selection $\omega_i$ as follows: the semantic score is the average closeness between instances in a selection that is:
$$S_{\mathcal T}(\omega_i) = \frac{\sum_{a \in \omega_i}{\sum_{b \in \omega_i,a\neq b }{CL_{\mathcal{T}}(a,b)}}}{|\omega_i|(|\omega_i|-1)}$$

We use $CL_{\mathcal{T}}(\cdot,\cdot)$ to denote the closeness between the classes of two samples according to the metadata $\mathcal{T}$ supplied by the hierarchical corpus. 
For example, the ImageNet consists of 1,000 classes of images organized in the form of a tree. In this tree, for two given nodes, we compute the distance, which is the length of the shortest path between them. The closeness of two instances in the selection is measured by using the inverse of distance of the shortest path between their respective nodes in the tree.
\input{figures/heatmap}
To see why this semantic score makes sense, in Figure~\ref{fig:heatmap}, we show the distance matrix of a small subset of ImageNet. In this small subset,  we can observe that for datasets semantically close to airplane, such as airliner, wing and warplane, distances of shortest paths among them are small. In contrast, their distances to the class shark is large.

\subsubsection{Remarks}\ \
Instead of treating the $n$-class ${\mathcal D}_{\omega}$ as a whole, the objectives function is broken down into $n$ parts, each corresponds to a class.   Computing the scores separately for each class is helpful when a model classifies very distinct objects with low semantic overlap. For example, a classifier may include cat as one class and airplane as another class.

The overall functional score is defined to be the sum of functional score of each class, that is:
   $$\mathcal{L}_{\mathcal{M},\mathcal{D}}({\omega}) = \sum_{1}^{n}{\mathcal{L}_{\mathcal{M}_i,\mathcal{D}}({\omega_i})}.$$
   
  Likewise, the semantic score is defined to be the sum of semantic score of each class, that is:
    $$S_{\mathcal T}(\omega) = \sum_{1}^{n}{S_{\mathcal T}({\omega}_i)}.$$

\subsection{Searching for Distributions}
We use a heuristic algorithm to search for the solution to our objective function. For each class $i$ of $\mathcal{M}$, we carry out the following steps:
\\ \\
{\bf Step 1: Compute the Functional Score.}\ \
Suppose there are $m$ leaf nodes in the corpus $\mathcal{C}$, we feed data of each leaf node ${\mathcal N}_j$ to $\mathcal{M}$ and compute  $l_{i,j}=\mathcal{L}_{\mathcal{M}_i,\mathcal{D}} ({\mathcal N}_j)$, for $j=1,\dots,m$.

For example, in CIFAR-10 dataset, the first class has label of `airplane'. When using the 1,000 leaf node datasets from ImageNet to compute the functional score, we can rank all datasets according to the score and get following list in Table~\ref{tab:relevance_rank}.
\input{figures/relevance_rank}
\noindent
{\bf Step 2: Compute Pairwise Closeness.}\ \
We select $l_{i,j}$ that is above a threshold. Let us write these $k$ largest as $l_{i,j_1}, l_{i,j_2}, ..., l_{i,j_k}$.  We then compute the pairwise  closeness, that is $CL_{\mathcal{T}}(j_a, j_b)$ for each $a,b$. 

Since datasets in ImageNet are organized in tree hierarchy, we are able to compute the distance of shortest path between any two nodes in the tree. Therefore, we define the closeness of two nodes as the inverse of the distance of shortest path between them.
\\ \\
\noindent
{\bf Step 3: Find a Subset.}\ \
From the pairwise closeness table, we can then use an algorithm to find a subset which can maximize the sum of functional score and semantic score.

In our implementation, we run K-medoids~\cite{DBLP:reference/ml/JinH10b} algorithm on the pairwise distance matrix. We also compute the distance between centers of clusters. If the distance is lower than a threshold $\eta$~\footnote{A larger $\eta$ is helpful in finding a more general input distribution. A smaller $\eta$ may find a more specific subclass but with some trade off in accuracy. In actual tasks, the investigators can try with different values until they get a reasonable result.}, we merge them into the same cluster. We then choose the cluster with the highest functional score.
\\ \\
\noindent
{\bf Step 4: Filtering.}\ \
Now we have a set of nodes (datasets). We would like to only keep the samples which can best represent the target class. For all samples in the chosen datasets, we only keep the data samples which get classified into the target class by the given target model.

%% file: figures/heatmap.tex
\begin{figure}[H]
    \centering
    \includegraphics[width=0.8\linewidth]{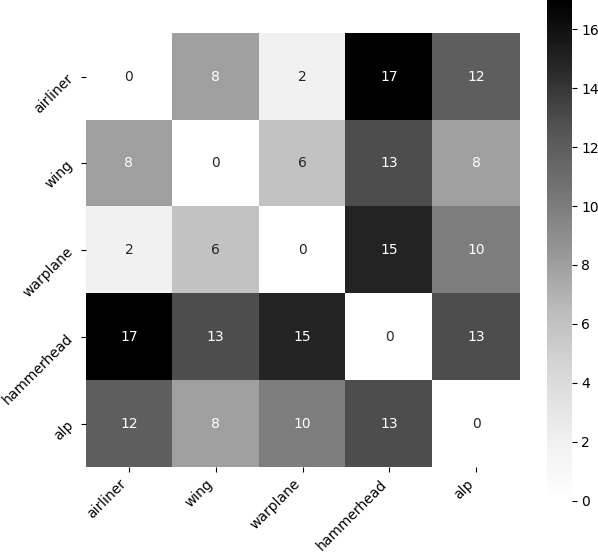}
    \caption{Heat-map of distance matrix among classes in a subset of ImageNet.}
    \label{fig:heatmap}
\end{figure}  

%% file: figures/relevance_rank.tex
\begin{table}[H]
\centering
\small
\begin{tabular}{|c|c|} 
\hline
\textbf{Dataset}          & \textbf{Score}  \\ 
\hline
airliner         & 0.94   \\ 
\hline
wing             & 0.92   \\ 
\hline
warplane         & 0.84   \\ 
\hline
hammerhead shark & 0.74   \\ 
\hline
alp              & 0.60    \\
\hline
...              & ...    \\
\hline
\end{tabular}
\caption{List of 1,000 classes ranked according to functional score regarding `airplane' class in CIFAR-10.}
\label{tab:relevance_rank}
\end{table}

%% file: eval.tex
\section{Evaluation}
\subsection{Setup}
Our evaluation uses five datasets: CIFAR-10~\cite{krizhevsky2009learning}, CIFAR-100~\cite{krizhevsky2009learning}, Caltech101~\cite{DBLP:journals/cviu/Fei-FeiFP07}, Oxford flowers~\cite{Nilsback08} and ImageNet~\cite{DBLP:conf/cvpr/DengDSLL009}. 
CIFAR-10 contains 50,000 training images and 10,000 test images. Caltech101 contains 8,677 images in total. Oxford flowers contains 1,020 training images and 6,149 test images. All samples are resized to $32 \times 32 \times 3$.

In our experimentation, CIFAR-10 and CIFAR-100 are mainly used to train classification models as targets for investigation. 

We then exploit the hierarchical information in ImageNet to investigate and find the data distribution of the target classes. The candidate data distribution we found is also used for further investigation: model cloning. 
These results of investigation are compared with investigation results with same procedure but using Caltech101 and Oxford flowers datasets.

\subsection{Evaluation on Functional Requirement}
In this section, we test the proposed approach when only functional requirement is applied.
\subsubsection{Controlled Environment Test on ImageNet}\ \
\label{sec:controlled}
We first test our method in a controlled environment where we randomly pick 10 classes from the whole 1,000 classes in ImageNet to train a target model. We chose simplified DLA~\cite{DBLP:conf/cvpr/YuWSD18} as the model architecture. We split the 12,850 images into 10,000 training and 2,850 testing images. This target model achieves 77.0\% accuracy.

We then compute the functional score between each class in this target model and each dataset in ImageNet independently using 50 testing images from each dataset. Note the images used in this computation are from testing set of ImageNet and have no overlap with images used in the training process of the target model.

In Table~\ref{tab:imagent_top2}, for each target class, we show the datasets with top-2 functional scores. We can observe that the dataset with the highest score is either the exact same class or a very similar class. In addition, the classes with second highest scores are also very close to the target classes.
\input{figures/imagenet_classifier}

\subsubsection{CIFAR-10 Classifier}\ \
\label{sec:cifar10}
We repeat the same experiment in Section~\ref{sec:controlled} on a dummy target classifier trained using CIFAR-10 dataset. 

We use the same simplified DLA architecture for the target model. For training the target model, we use 45,000 out of 50,000 images in the training dataset of CIFAR-10. This target model achieves 94.9\% accuracy.

The remaining 5,000 training images are reserved for the model cloning performance comparison later in Section~\ref{sec:cloning}.

In Table~\ref{tab:cifar10_top2}, we can observe the top-1 and top-2 classes are mostly very close to the original class with one exception in the class `truck'. This is probably caused by the visual similarity (rectangular shape) between truck and entertainment center.
\input{figures/cifar10_classifier}

\subsection{Evaluation on Semantic Requirement}

Section~\ref{sec:controlled} shows that the functional score term of objective function works well. In this section, we proceed to test the full objective function with semantic score included. We also use the proposed search algorithm to find the candidate dataset. 

Here we set the target classifier as the CIFAR-10 classifier we used in Section~\ref{sec:cifar10}. In Table~\ref{tab:w2v}, we list the labels of first five classes in CIFAR-10 in the first column. In the second column, we list the names of the datasets which we pick samples from. In the third column, we show the functional scores of the datasets selected by the proposed method, with respect to each class in the target model.
\newpage
\noindent
{\bf Verify the Results using Label Similarity.}\ \
\label{sec:verify}
The model predicted a union of classes, and we know the ground truth of each class. However, verifying whether the found distribution is the correct one is not straightforward. 

The main challenge is that different datasets may use different terms to refer to a same class, so it is not straightforward to compare two labels and decide whether they are close or distant. For example, how to verify whether the union of classes `airliner, wing, warplane, military' is a good representation of the ground truth `airplane' is not obvious.
\\ \\
\noindent
{\em Word Embedding.}\ \
Word embedding is the representation of words for text analysis, typically in the form of a real-valued vector that encodes the meaning of the word such that the words that are closer in the vector space are expected to be similar in meaning. Word embeddings can be obtained using a set of language modeling and feature learning techniques where words or phrases from the vocabulary are mapped to vectors of real numbers. There are many existing approaches to generate word embedding, for example, word2vec~\cite{mikolov2013efficient} and GloVe~\cite{pennington-etal-2014-glove}.

As different datasets may use different words/phrases to label the same data distribution, word embedding~\cite{mikolov2013efficient,pennington-etal-2014-glove} can be useful in checking whether they are referring to the same thing. For example, for the class airplane in CIFAR-10, we have similar classes such as airliner, warplane and wing in ImageNet. 

Suppose we can find a candidate distribution of a target class in the unknown model, we can compare the label of this selected distribution and the label of that class in the original training dataset to check whether they are the same or how similar they are. The comparison can be done by measuring the euclidean distance between their word embeddings.
\input{figures/word2vec}
We use a pre-trained Word2Vec~\cite{mikolov2013efficient} embedder to compute the cosine similarity between embeddings of labels. For labels which are phrases of multiple tokens, we set the similarity between two phrases as the similarity between their nearest tokens.

We show this similarity score in the last column in Table~\ref{tab:w2v}. The results show that the similarities between names of found datasets and their corresponding ground truth labels are high. For a candidate distribution with phrases containing the original class name, we can confirm the selection is good with high confidence. For example, the predicted union of classes for the original class `cat' is good as it contains all specific breeds of cats.

In Figure~\ref{fig:tsne}, we visualize word embeddings of the labels found by the proposed method and ground truths by mapping them to two-dimensional space using t-distributed stochastic neighbor embedding (t-SNE). We can see the labels found by the proposed method are indeed close to their corresponding ground truths in the embedding space.
\input{figures/tsne}

\subsection{Further Investigation: Model Cloning}
\label{sec:cloning}
To further evaluate the adequacy of the datasets found by the proposed method, we conduct model cloning by using the found datasets as auxiliary datasets. We compare the performance of cloning using: (1) {Same Distribution}: clone using the same CIFAR distribution (split data); (2) {Similar Distribution}: clone using Caltech101 dataset; (3) {Distant Distribution}: clone using Oxford flowers dataset. (4) Clone using randomly sampled subset from ImageNet.

\subsubsection{Target Model: CIFAR-10 Classifier}\ \
We first use the target model (with simplified DLA architecture) in Section~\ref{sec:cifar10}. As this model was using 45,000 training images of CIFAR-10, we use remaining 5,000 training images to conduct model cloning. The accuracy of the target model is shown in Table~\ref{tab:search_clone_accuracy} column 2.

We use GoogLeNet~\cite{DBLP:conf/cvpr/SzegedyLJSRAEVR15} as the cloned model's architecture and train each clone for 200 epochs. Here we follow Papernot \etal's method of cloning~\cite{DBLP:conf/ccs/PapernotMGJCS17}. The accuracy for each class and overall accuracy is shown in Table~\ref{tab:search_clone_accuracy} column 3.
We then repeat the same cloning procedure for two different dataset: Caltech101 (column 4) and Oxford flowers (column 5). We also only use 5,000 images from each dataset and use the same GoogLeNet architecture. 
For comparison, we also randomly sample 5,000 images from ImageNet without using the search algorithm to conduct cloning (column 6).

We then conduct the same model cloning process for datasets found by the proposed method. For fair comparison, the dataset found by proposed method also contains 5,000 images in total.
Column 7 shows the cloning performance using the datasets we found based on the first term (functional score) of the objective function. Column 8 shows the performance when both functional score and semantic score are considered. 

In Table~\ref{tab:search_clone_accuracy}, we can observe that cloning using the same distribution gives the highest accuracy. Caltech101 contains many similar classes as the target model, for example, passenger vehicles and animals, thus has a relatively high cloning accuracy. In contrast, Oxford flowers only contains images of flowers, therefore results in the lowest cloning accuracy.

With only one term of the objective function enabled, the performance of the proposed method has already surpassed randomly sampling data from ImageNet. 
We can observe that the performance is better with both terms enabled since the obtained datasets are more semantically meaningful.

\input{figures/other_methods}

\subsubsection{Target Model: CIFAR-100 Classifier}\ \
We repeat the same experiment of model cloning on a target model with same architecture but trained using CIFAR-100. The model was also trained using 45,000 images and 5,000 images were reserved for testing the cloning performance from same distribution. This target model achieves 76.4\% accuracy. 

As there are 100 classes in CIFAR-100, we are unable to show a full table of accuracy here. We show the KDE (Kernel density estimation) plot instead. The horizontal axis shows the accuracy and the vertical axis shows the density.

In Figure~\ref{fig:cifar-100}, we can observe that the result is similar to what we have seen on CIFAR-10. The dataset obtained through the proposed method yields much better cloning accuracy than a hand-picked distribution that is seemingly close to the training distribution.
\input{figures/kde.tex}
\subsection{Field Experiment: Hugging Face Model Hub}
\label{sec:field}
The Hugging Face platform stores over 60,000 machine learning models. These model are mostly open source and can be directly downloaded. 

While most models have detailed documentations, some only provide the code to load the checkpoints without any description of the purpose/usage of the model. In this section, we randomly pick two undocumented models and use the proposed method to investigate them. There is no documentation of the models in these two repositories at the time of submission of this manuscript.
\\ \\
\noindent
{\bf \em First Example:}\\
{\url{https://huggingface.co/Matthijs/mobilenet_v2_0.75_160}}.\\
This model has 1,000 classes. There seems to be a resemblance between the training data distribution of this model and ImageNet. For 991 out of 1,000 classes, the search algorithm converges to a leaf node in ImageNet with average functional score of 0.77. In addition, more than 75\% of the classes have their functional scores higher than 0.9. Further analysis shows that the indices of classes in this unknown model also match well with the class indices of leaf nodes found by the search algorithm. When we treat the unknown model as an ImageNet classifier, it can achieve 69.5\% accuracy on the testing dataset and 99.5\% accuracy on the training dataset.
Therefore, based on these evidences, we highly suspect that this model is trained using ImageNet dataset.
\\ \\
\noindent
{\bf \em Second Example:}\\
{\url{https://huggingface.co/nateraw/timm-resnet50-beans-copy}}.\\
This model has three classes.
The search algorithm matches all three classes to node n00017222 (plant, flora, plant life) in ImageNet with functional score of 0.81, 0.74, 0.72 respectively.
We visually inspected node n00017222 and found out that n00017222 consists of two classes: yellow lady's slipper and daisy. Most images contain both flowers and leaves.
As the node has quite limited number of images, in our attempt to find out the exact classes of the unknown classifier, we apply two other datasets in our investigation: Oxford flowers~\cite{Nilsback08} and Flavia leaf dataset~\cite{DBLP:journals/corr/abs-0707-4289}.

As both Oxford flowers and Flavia leaf dataset are not hierarchical datasets, here our analysis relies on the functional score only. When using Oxford flowers, the functional scores are generally low with an average of only 0.36. On the other hand, on Flavia leaf dataset, the average functional score is 0.83, which is even higher than the average score of node n00017222.

Flavia leaf dataset contains 1,907 images from 33 classes of leaves. Among all the classes, the class `Cercis chinensis' has the highest average functional score of 0.90. 

\begin{figure}[H]
    \centering
	\begin{subfigure}{.25\linewidth}
		\centering
		\includegraphics[width=1\linewidth]{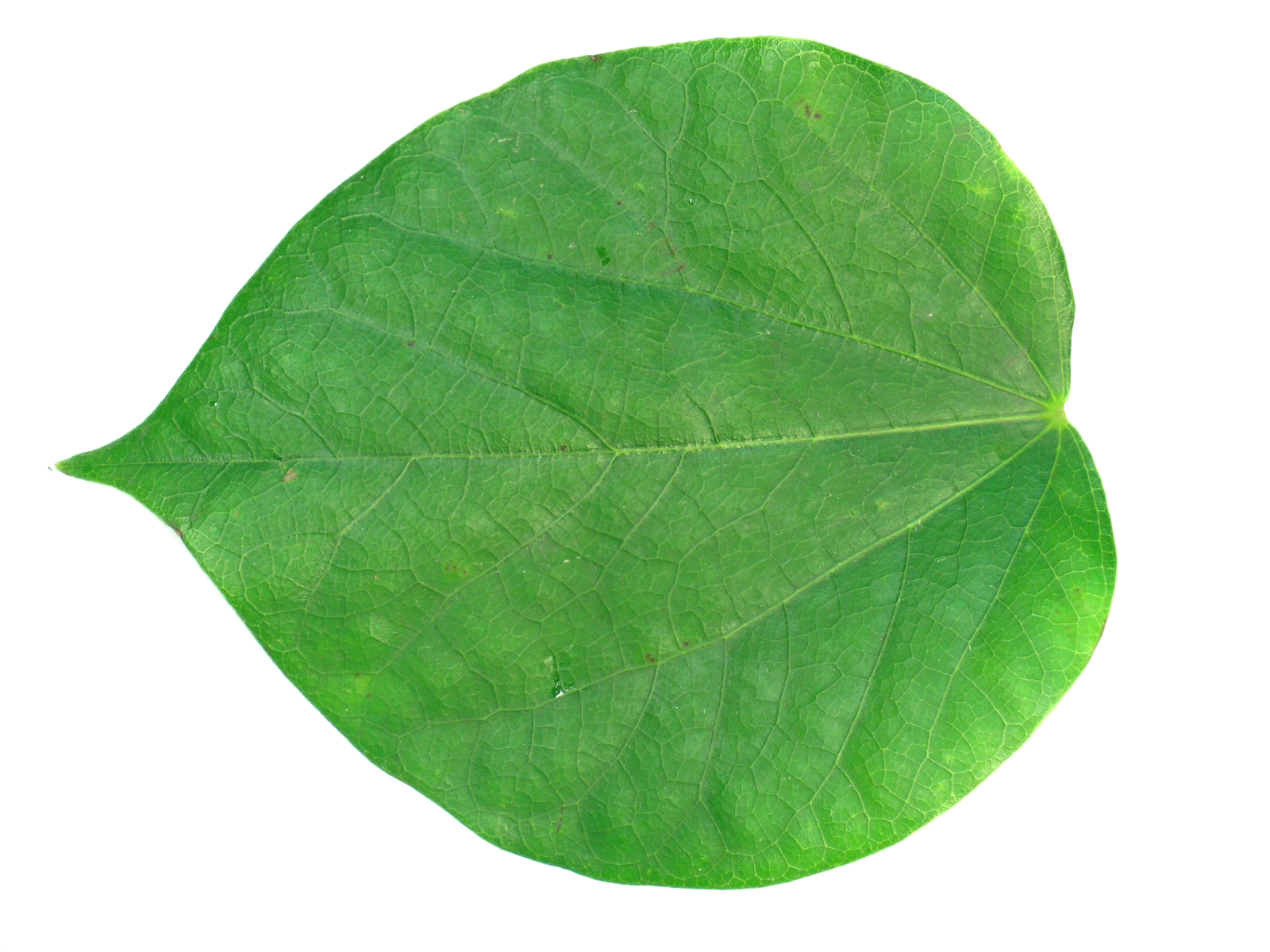}
	  \end{subfigure}%
	  \begin{subfigure}{.25\linewidth}
		\centering
		\includegraphics[width=1\linewidth]{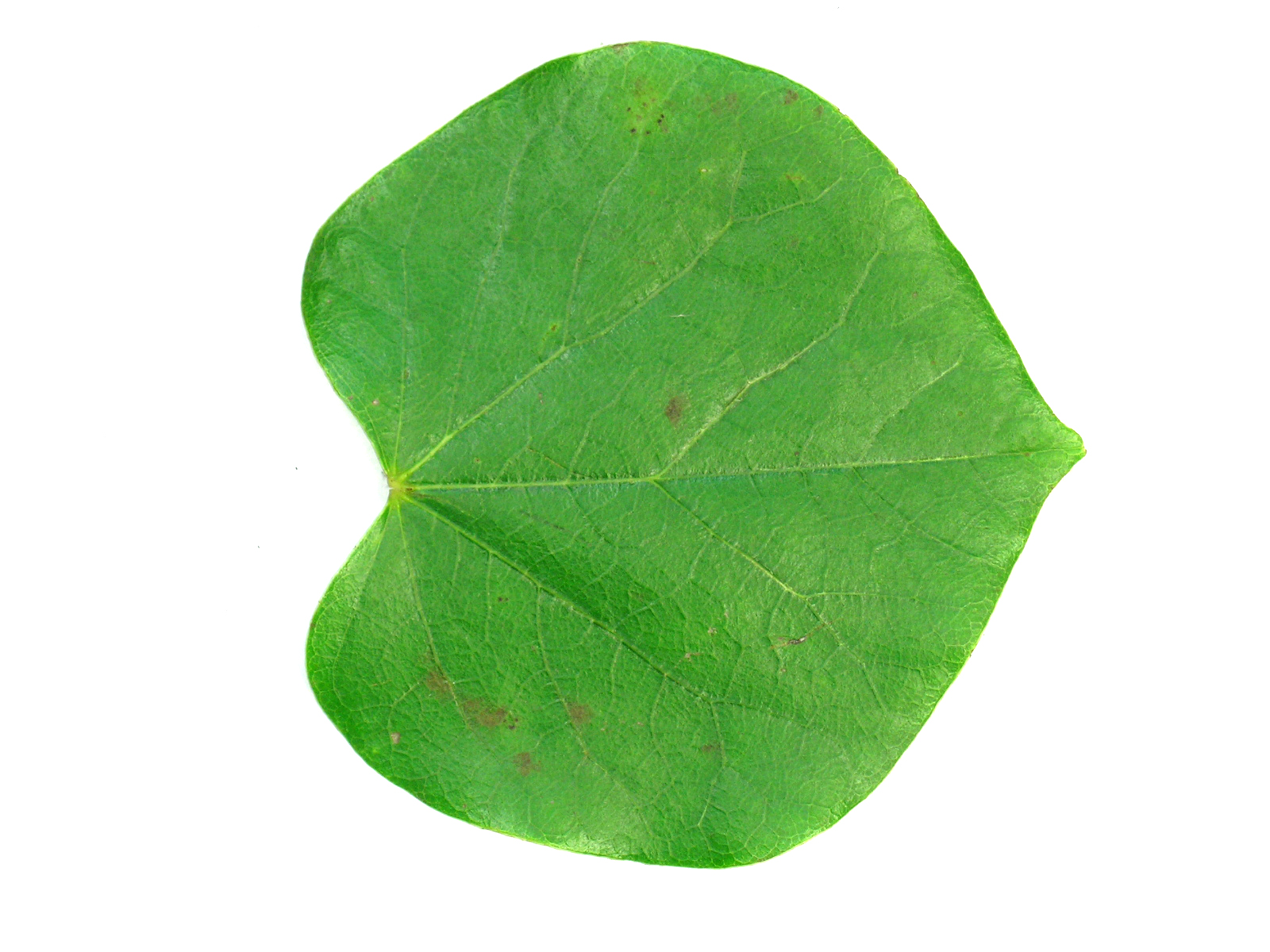}
	  \end{subfigure}%
	  \begin{subfigure}{.25\linewidth}
		\centering
		\includegraphics[width=1\linewidth]{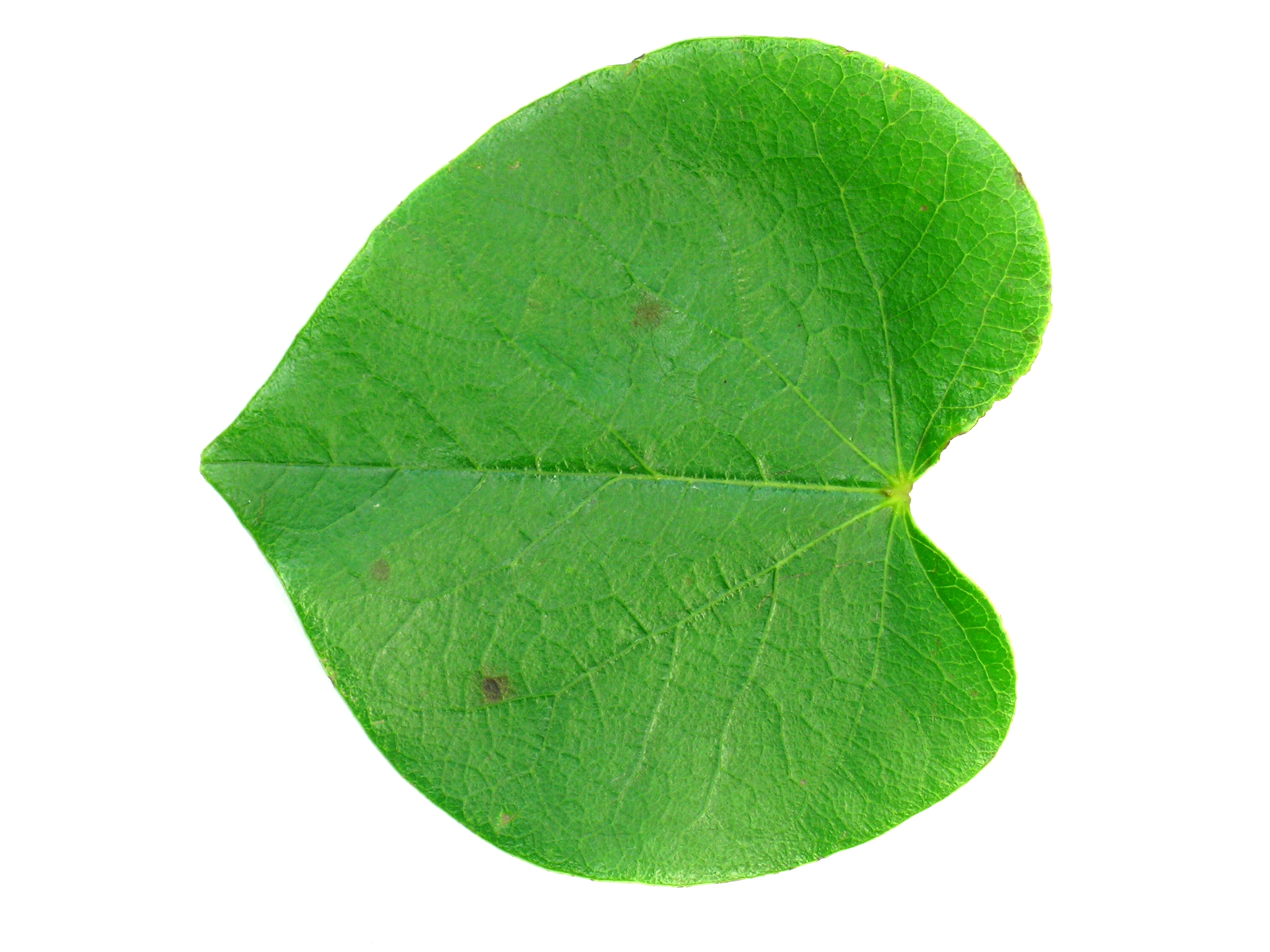}
	  \end{subfigure}%
	  \begin{subfigure}{.25\linewidth}
		\centering
		\includegraphics[width=1\linewidth]{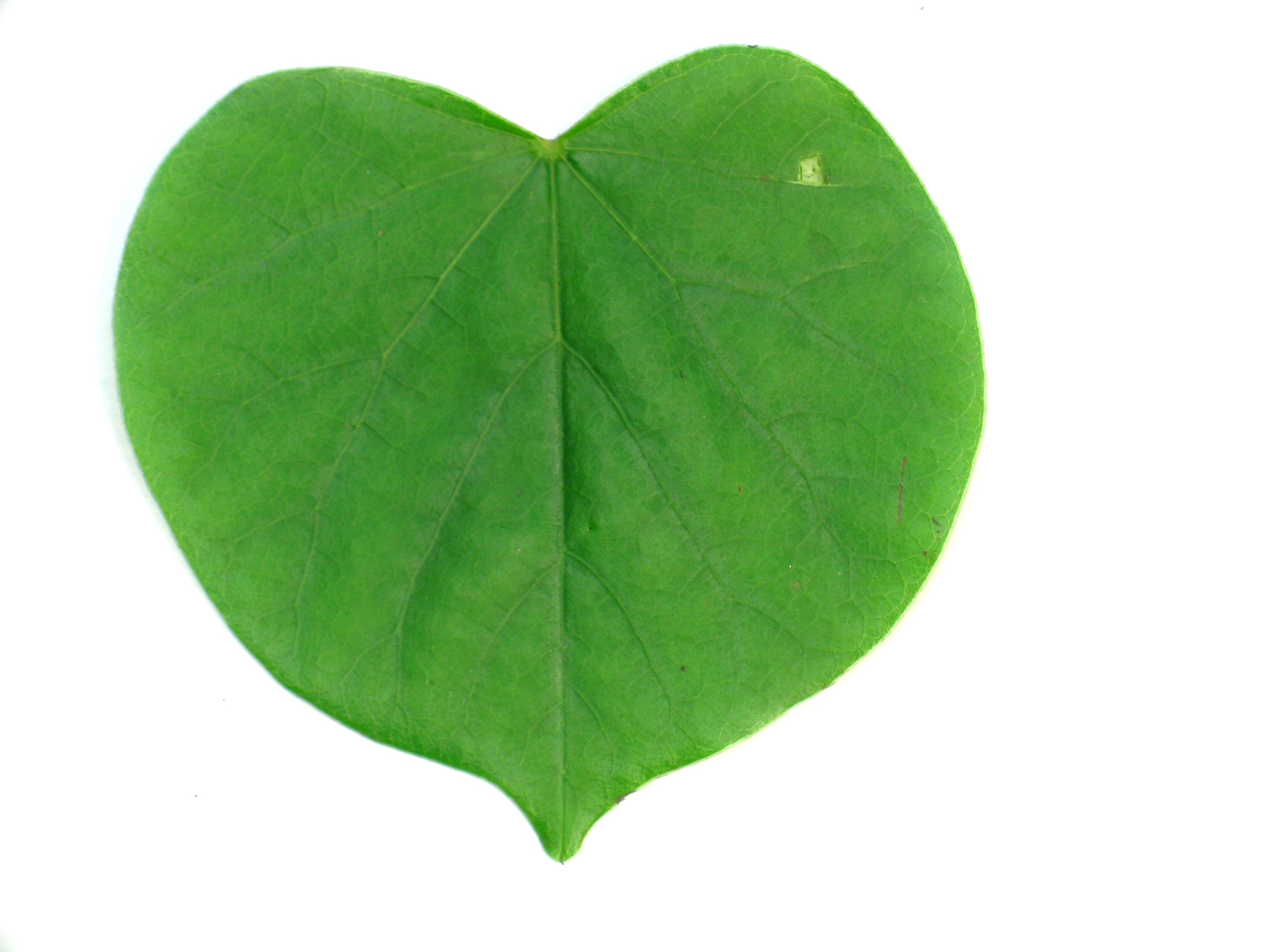}
	  \end{subfigure}
	  \caption{Four samples with the highest functional scores in Flavia leaf dataset.}
      \label{fig:leaves}
\end{figure}

In Figure~\ref{fig:leaves}, we show four samples with the highest functional scores for the first class of the unknown model (left is the highest).
Cercis chinensis is a plant from the bean family. It is possible that the unknown model is a classifier for some leaves visually similar to bean leaves.

%% file: figures/imagenet_classifier.tex
\begin{table}[H]
\centering
\small
\setlength{\tabcolsep}{1pt}
\begin{tabular}{|c|c|c|c|c|} 
    \hline
    \textbf{ Original Class } & \textbf{Top 1 }    & \textbf{Score } & \textbf{Top 2 }       & \textbf{Score }  \\ 
    \hline
    kit fox                   & kit fox            & 0.82            & yellow lady's slipper & 0.70             \\ 
    \hline
    English setter            & clumber spaniel    & 0.86            & dalmatian             & 0.58             \\ 
    \hline
    Siberian husky            & Eskimo dog, husky  & 0.74            & malamute              & 0.68             \\ 
    \hline
    Australian terrier        & Australian terrier & 0.74            & silky terrier         & 0.68             \\ 
    \hline
    English springer          & English springer   & 0.84            & Bernese mountain dog  & 0.56             \\ 
    \hline
    grey whale                & dugong             & 0.94            & tiger shark           & 0.90             \\ 
    \hline
    lesser panda              & lesser panda       & 0.92            & ladybug               & 0.54             \\ 
    \hline
    Egyptian cat              & tabby cat          & 0.64            & leopard               & 0.60             \\ 
    \hline
    ibex                      & ibex               & 0.88            & bighorn               & 0.60             \\ 
    \hline
    Persian cat               & Persian cat        & 0.78            & mosquito net          & 0.48             \\
    \hline
    \end{tabular}
\caption{Top 2 classes ranked by functional scores for each target class in a classier trained using 10 classes from ImageNet.}
\label{tab:imagent_top2}
\end{table}

%% file: figures/cifar10_classifier.tex
\begin{table}[H]
\centering
\small
\setlength{\tabcolsep}{1pt}
\begin{tabular}{|c|c|c|c|c|} 
    \hline
    \begin{tabular}[c]{@{}c@{}}\textbf{ Original}\\\textbf{Class }\end{tabular} & \textbf{Top 1 }      & \textbf{Score } & \textbf{Top 2 }   & \textbf{Score }  \\ 
    \hline
    airplane                                                                    & airliner             & 0.94            & wing              & 0.92             \\ 
    \hline
    automobile                                                                  & convertible          & 0.84            & sports car        & 0.84             \\ 
    \hline
    bird                                                                        & blue heron           & 0.88            & jay               & 0.86             \\ 
    \hline
    cat                                                                         & tabby cat            & 0.92            & Persian cat       & 0.92             \\ 
    \hline
    deer                                                                        & hartebeest           & 0.86            & impala            & 0.78             \\ 
    \hline
    dog                                                                         & Japanese spaniel     & 0.94            & Dandie Dinmont    & 0.92             \\ 
    \hline
    frog                                                                        & bullfrog             & 0.74            & hen of the woods  & 0.70             \\ 
    \hline
    horse                                                                       & sorrel               & 0.84            & football helmet   & 0.54             \\ 
    \hline
    ship                                                                        & schooner             & 0.96            & drilling platform & 0.94             \\ 
    \hline
    truck                                                                       & entertainment center & 0.96            & forklift          & 0.90             \\
    \hline
    \end{tabular}
\caption{Top 2 classes ranked by functional scores for each target class in a classifier trained using CIFAR-10 images.}
\label{tab:cifar10_top2}
\end{table}

%% file: figures/word2vec.tex
\vspace{-5pt}
\begin{table}[H]
    \small
    \centering
    \setlength{\tabcolsep}{1pt}
    \begin{tabular}{|c|c|c|c|} 
        \hline
        \begin{tabular}[c]{@{}c@{}}\textbf{ Original}\\\textbf{Class }\end{tabular} & \textbf{Predicted~Union of~Classes }                                                                                                             & \begin{tabular}[c]{@{}c@{}}\textbf{Func}\\\textbf{-tional}\\\textbf{Score }\end{tabular} & \begin{tabular}[c]{@{}c@{}}\textbf{Word}\\\textbf{-2Vec}\\\textbf{Score }\end{tabular}  \\ 
        \hline
        airplane                                                                    & airliner, wing, warplane, military plane                                                                                                         & 0.90                                                                                     & 0.581                                                                                   \\ 
        \hline
        automobile                                                                  & \begin{tabular}[c]{@{}c@{}}convertible, sports car, minivan, beach \\wagon, station wagon, wagon, estate car\end{tabular}                        & 0.78                                                                                     & 0.394                                                                                   \\ 
        \hline
        bird                                                                        & \begin{tabular}[c]{@{}c@{}}little blue heron, Egretta, caerulea,\\jay, jacamar, magpie, junco,\\snowbird, ostrich, Struthio camelus\end{tabular} & 0.75                                                                                     & 0.512                                                                                   \\ 
        \hline
        cat                                                                         & \begin{tabular}[c]{@{}c@{}}tabby cat, Persian cat, Siamese \\cat, Egyptian cat, tiger cat\end{tabular}                           & 0.81                                                                                     & 1.000                                                                                   \\ 
        \hline
        deer                                                                        & \begin{tabular}[c]{@{}c@{}}hartebeest, impala, Aepyceros\\melampus, gazelle\end{tabular}                                                         & 0.74                                                                                     & 0.412                                                                                   \\
        \hline
        \end{tabular}
\caption{Functional score and word2vec cosine similarity of resultant datasets. (w.r.t. first 5 classes in CIFAR-10 target model.)}
\label{tab:w2v}
\end{table}
\vspace{-15pt}

%% file: figures/tsne.tex
\begin{figure}[H]
    \centering
    \includegraphics[width=\linewidth]{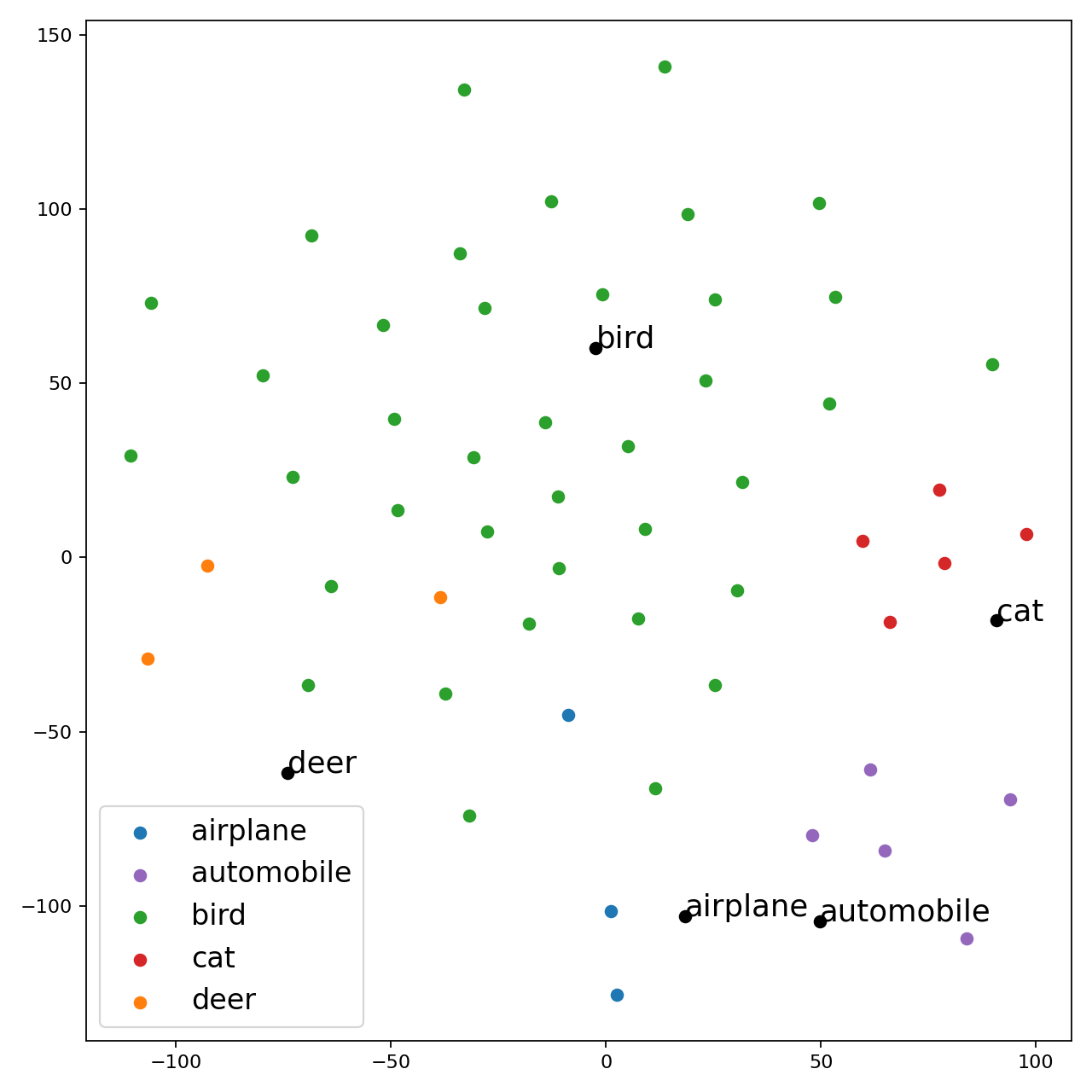}
    \caption{t-SNE visualization of embeddings of ground truths labels and labels obtained through the proposed method.}
    \label{fig:tsne}
\end{figure}

%% file: figures/other_methods.tex
\begin{table}[H]
    \small
    \centering
    \setlength{\tabcolsep}{1pt}
    \begin{tabular}{|c|c|c|c|c|c|c|c|} 
        \hline
        \begin{tabular}[c]{@{}c@{}}\\\textbf{ Class }\end{tabular} & \begin{tabular}[c]{@{}c@{}}\textbf{Target}\\\textbf{Model}\end{tabular} & \begin{tabular}[c]{@{}c@{}}\textbf{CIFAR}\\\textbf{10}\end{tabular} & \begin{tabular}[c]{@{}c@{}}\textbf{Caltech}\\\textbf{101}\end{tabular} & \begin{tabular}[c]{@{}c@{}}\textbf{Oxford}\\\textbf{Flowers}\end{tabular} & \begin{tabular}[c]{@{}c@{}}\textbf{Image}\\\textbf{-Net}\end{tabular} & \begin{tabular}[c]{@{}c@{}}\textbf{Extract}\\\textbf{using}\\\textbf{1 term}\end{tabular} & \begin{tabular}[c]{@{}c@{}}\textbf{Extract}\\\textbf{using}\\\textbf{2 terms}\end{tabular}  \\ 
        \hline
        airplane                                                   & 95.3\%                                                                  & 84.5\%                                                              & 73.7\%                                                                 & 72.7\%                                                                    & 35.8\%                                                                & 78.9\%                                                                                      & 76.6\%                                                                                        \\ 
        \hline
        automobile                                                 & 97.8\%                                                                  & 90.7\%                                                              & 0.1\%                                                                  & 0.0\%                                                                     & 44.6\%                                                                & 94.1\%                                                                                      & 92.9\%                                                                                        \\ 
        \hline
        bird                                                       & 92.8\%                                                                  & 68.1\%                                                              & 17.2\%                                                                 & 24.7\%                                                                    & 15.2\%                                                                & 39.1\%                                                                                      & 43.3\%                                                                                        \\ 
        \hline
        cat                                                        & 88.6\%                                                                  & 74.7\%                                                              & 77.1\%                                                                 & 68.1\%                                                                    & 41.3\%                                                                & 71.2\%                                                                                      & 74.4\%                                                                                        \\ 
        \hline
        deer                                                       & 96.6\%                                                                  & 80.8\%                                                              & 72.4\%                                                                 & 49.7\%                                                                    & 46.5\%                                                                & 61.2\%                                                                                      & 70.3\%                                                                                        \\ 
        \hline
        dog                                                        & 91.9\%                                                                  & 69.2\%                                                              & 3.4\%                                                                  & 4.0\%                                                                     & 54.0\%                                                                & 61.4\%                                                                                      & 77.7\%                                                                                        \\ 
        \hline
        frog                                                       & 97.4\%                                                                  & 86.2\%                                                              & 72.7\%                                                                 & 65.4\%                                                                    & 20.1\%                                                                & 64.4\%                                                                                      & 78.5\%                                                                                        \\ 
        \hline
        horse                                                      & 96.5\%                                                                  & 78.5\%                                                              & 0.3\%                                                                  & 0.0\%                                                                     & 31.1\%                                                                & 53.9\%                                                                                      & 57.8\%                                                                                        \\ 
        \hline
        ship                                                       & 96.0\%                                                                  & 90.7\%                                                              & 83.8\%                                                                 & 59.6\%                                                                    & 43.2\%                                                                & 67.4\%                                                                                      & 68.9\%                                                                                        \\ 
        \hline
        truck                                                      & 96.4\%                                                                  & 89.8\%                                                              & 54.3\%                                                                 & 0.0\%                                                                     & 23.2\%                                                                & 38.7\%                                                                                      & 79.8\%                                                                                        \\ 
        \hline
        \textbf{Average }                                          & \textbf{94.9\%}                                                         & \textbf{81.3\%}                                                     & \textbf{45.5\%}                                                        & \textbf{34.4\%}                                                           & \textbf{35.5\%}                                                       & \textbf{\textbf{63.0\%}}                                                                    & \textbf{72.0\%}                                                                               \\
        \hline
        \end{tabular}
    \caption{Comparison of accuracy of models cloned by using various given datasets and models cloned by using datasets extracted with the proposed method.}
    \label{tab:search_clone_accuracy}
    \end{table}

%% file: figures/kde.tex
\begin{figure}[H]
\centering
\begin{subfigure}{0.5\linewidth}
\centering
\vspace{0pt} \includegraphics[width=\linewidth]{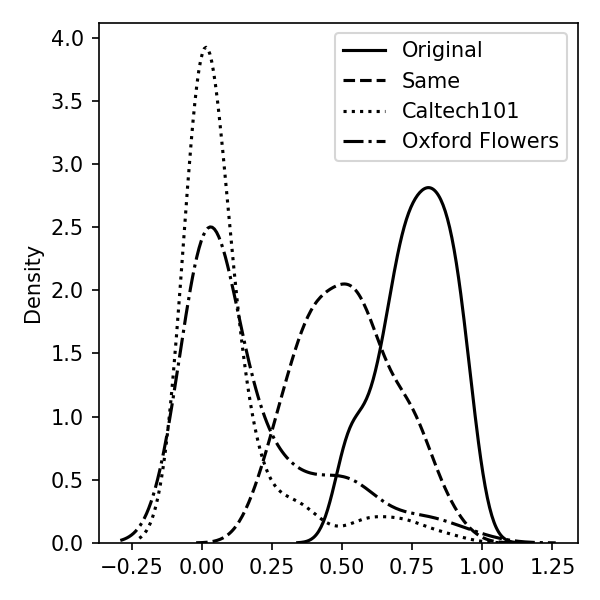}
\caption{Original model vs. \\ model cloned using CIFAR-100,\\Caltech101 and Oxford Flowers\\datasets.}
\label{fig:cosine}
\end{subfigure}%
\begin{subfigure}{0.5\linewidth}
\centering
 \includegraphics[width=\linewidth]{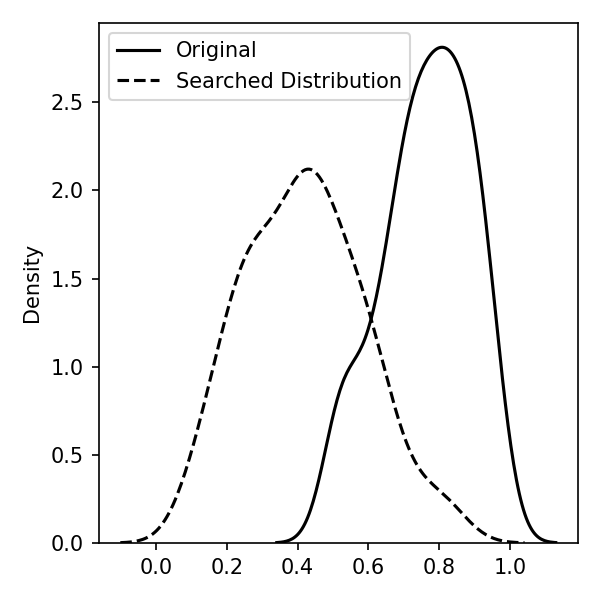}
\caption{Original model vs.\\ model cloned using datasets\\obtained through the proposed\\approach.}
\label{fig:density}
\end{subfigure}
\caption{KDE plot of accuracy of each class (100 classes).}
\label{fig:cifar-100}
\end{figure}

%% file: countermeasure.tex
\section{Future Works}
\subsection{Requirement on Hierarchical Datasets}
The proposed method is able to find distributions that are close to the training distribution of a black-box model if such close distributions exist in the given hierarchical corpus.
When running the algorithm for a model with no similar distribution in the corpus, it returns a `not found' result.

Constructing a more comprehensive hierarchical corpus relies on community effort. Over the past few years, we have seen large datasets such as Google's JFT-3B~\cite{DBLP:journals/corr/abs-2106-04560} which was annotated with a class-hierarchy of around 30k labels. There are also large hierarchical dataset for specific domain such as Stanford Medical ImageNet~\cite{medical}. However, these datasets are not publicly available yet.

While ImageNet is far from being comprehensive, it is a good starting point. As a hierarchical corpus, it is very expandable. For each specific investigation task, we can add our data to a node in the ImageNet hierarchy. Based on our experience in Section~\ref{sec:field}, the process is straightforward and the outcome is promising. 

\begin{figure}[H]
    \centering
	\begin{subfigure}{.25\linewidth}
		\centering
		\includegraphics[width=1\linewidth]{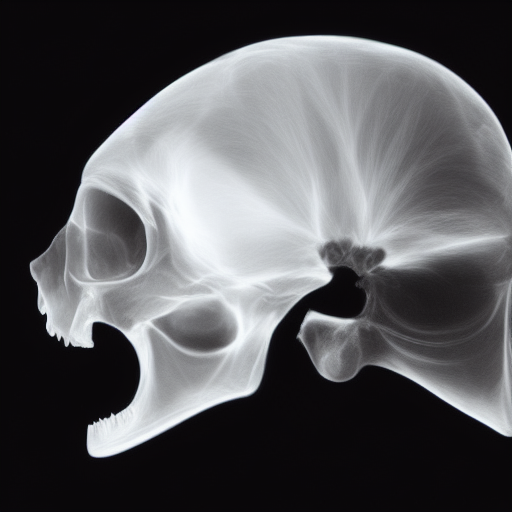}
		\caption{Cat X-ray.}
	  \end{subfigure}%
	  \begin{subfigure}{.25\linewidth}
		\centering
		\includegraphics[width=1\linewidth]{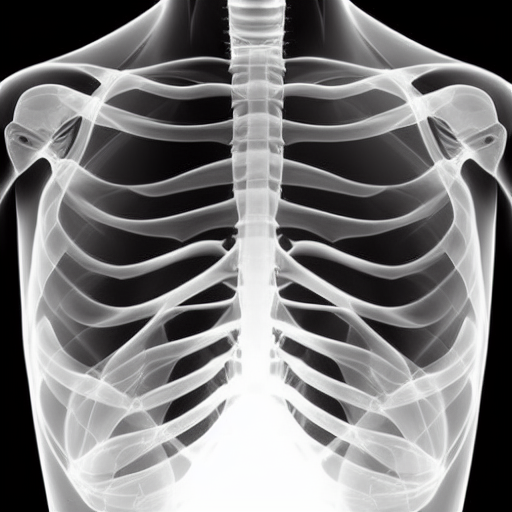}
		\caption{Chest X-ray.}
	  \end{subfigure}%
	  \begin{subfigure}{.25\linewidth}
		\centering
		\includegraphics[width=1\linewidth]{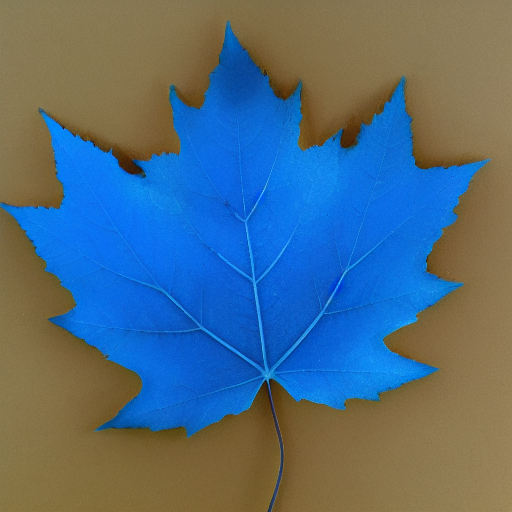}
		\caption{Blue maple.}
	  \end{subfigure}%
	  \begin{subfigure}{.25\linewidth}
		\centering
		\includegraphics[width=1\linewidth]{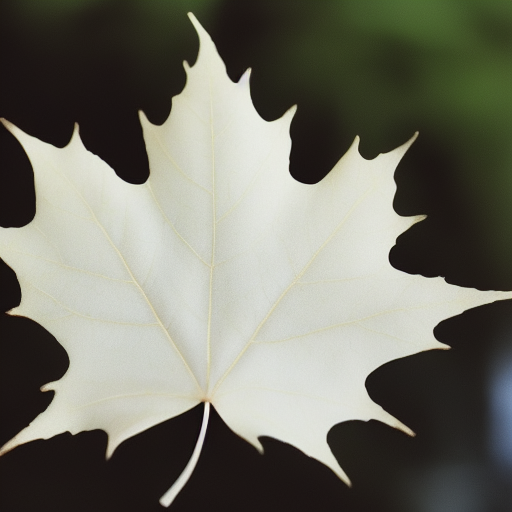}
		\caption{White maple.}
	  \end{subfigure}

	  \caption{Examples of generated samples in new subtrees.}
      \label{fig:diffusion}
\end{figure}

In addition, with recent breakthrough in generative models, it becomes possible to generate images to expand a given corpus.
For example, one limitation of ImageNet is the lack of X-ray images. We can use stable diffusion to generate a large set of X-ray images (shown in Figure~\ref{fig:diffusion}), organized them in the form of a tree and add them to ImageNet as a subtree. Although they may contain errors sometimes, most images are good enough for analysis under the proposed method. Similarly, we can also generate new data for `leaf', even for samples like blue maple and white maple which do not exist in real life (shown in Figure~\ref{fig:diffusion}).

When a candidate distribution is extracted from a hierarchical corpus, the images are guaranteed to be meaningful natural images.
However, when using generative models, additional constraints may be required to make sure the selected datasets contain semantically meaningful objects. 
Certain distributions of noises, artifacts or adversarial images may fulfill functional requirements but will not be useful for forensic investigation. 

In Figure~\ref{fig:diffusion}, the process of expanding the hierarchical corpus was done manually. An automatic workflow can also be easily implemented. Testing samples can be generated `on the fly' with prompts progressively narrow down from a general concept to a specific object. The research on such algorithm will be an interesting future work.

Another important future work will be applying the proposed method to unknown models trained using other data types such as text and graph.
Our objective function and search method are compatible with different data types since we use a model's hard label output regardless of the internal architecture. Theoretically, changing to a hierarchical corpus of other data type will work without additional modification.

\subsection{Use as an Attack and Potential Countermeasures}
The proposed method is a powerful tool for forensic investigation. However, it may be used for attacks as well. While there are some relevant works on potential defenses, they are mostly for data protection during federated learning~\cite{DBLP:conf/icml/Huang0LA20} or running on untrusted hardware~\cite{DBLP:conf/icml/AlamROH22}.

Since the attack is new, we designed a custom countermeasure which may potentially work. 
We propose hiding the real input distribution of a model by adaptively planting some back-doors which generate strong signals that confuse the investigator/attacker. 

We trained a MNIST~\cite{lecun2010mnist} classifier to verify our hypothesis. The model is trained in such way: We use the usual 60,000 training images; We pair the 10 classes in MNIST with the 10 classes in CIFAR-10; For each class in MNIST, we add 500 additional training images from their respective CIFAR-10 class to inject the back-doors. We use simplified DLA as the architecture. The model was trained for 200 epochs and reached 99.2\% accuracy. 

When running the proposed search algorithm on this classifier using ImageNet as the corpus, it returns datasets which are obviously more related to the CIFAR-10 classes. For example, the digit $0$ was paired with the airplane class during training. The search result for this class was `wing, airliner, warplane and airship' with a high functional score at 0.745.

%% file: conclusion.tex
\section{Conclusion}
We highlight that knowing the data domain of a model is a necessary first step before conducting further investigation. We present a method that allows us to extract a meaningful distribution that is similar to the model's input data distribution from a hierarchical dataset. The experimentation results show that the proposed method is highly useful and very accurate even with only black-box and hard-label only access to the target model being investigated. In addition, when using the dataset obtained through the proposed method to conduct model cloning, the performance is much better than manually choosing a similar distribution to clone. We also discussed potential countermeasures which could be employed to hinder the proposed investigation method.